\documentclass{article}
\usepackage{ijcai19}

\usepackage{times}
\usepackage{soul}
\usepackage{url}
\usepackage[hidelinks]{hyperref}
\usepackage[utf8]{inputenc}
\usepackage[small]{caption}
\usepackage{graphicx}
\usepackage{amsmath}
\usepackage{booktabs}
\usepackage{amsfonts}
\usepackage{nicefrac}
\usepackage{algorithm}
\usepackage{algorithmic}
\graphicspath{{figures/}}
\newtheorem{problem}{Problem}
\DeclareUnicodeCharacter{00A0}{ }

\title{Semi-supervised Learning on Graph with an Alternating Diffusion Process}

\author{
Qilin Li\footnote{Contact Author}\and
Senjian An\and
Ling Li\And
Wanquan Liu\\
\affiliations
Department of Computing, Curtin University, Perth, Australia \\
\emails
li.qilin@postgrad.curtin.edu.au,
\{s.an, l.li, w.liu\}@curtin.edu.au
}

\begin{document}

\maketitle

\begin{abstract}
Graph-based semi-supervised learning usually involves two separate stages, constructing an affinity graph and then propagating labels for transductive inference on the graph. It is suboptimal to solve them independently, as the correlation between the affinity graph and labels are not fully exploited. In this paper, we integrate the two stages into one unified  framework by formulating the graph construction as a regularized function estimation problem similar to label propagation. We propose an alternating diffusion process to solve the two problems simultaneously, which allows us to learn the graph and unknown labels in an iterative fasion. With the proposed framework, we are able to adequately leverage both the given labels and estimated labels to construct a better graph, and effectively propagate labels on such a dynamic graph updated simultaneously with the newly obtained labels. Extensive experiments on various real-world datasets have demonstrated the superiority of the proposed method compared to other state-of-the-art methods.       
\end{abstract}

\section{Introduction}
Semi-supervised learning (SSL) refers to the problem of learning from labeled and unlabeled data. SSL is of particular interest in many real-world applications since labeled data is often scarce, whereas unlabeled data is abundant. One family of SSL methods known as graph-based approaches have attracted much attention due to their elegant formulation and high performance. 

Graph-based semi-supervised learning (GSSL) represents both labeled and unlabeled data as vertices in a undirected graph $G=(V,E)$, where edges between vertices are weighted by the corresponding pair-wise affinities/similarities. The key to GSSL is the manifold/cluster assumption saying that points on the same manifold are likely to have the same label~\cite{zhou2004learning}. With the large portion of unlabeled points revealing the underlying manifold structure represented by the graph, the small portion of labeled points are then used to perform label propagation for transductive inference on unlabeled points. 

Over the last decade, many works of GSSL, such as Label Propagation~\cite{zhu2002learning}, GFHF~\cite{zhu2003semi} and LGC~\cite{zhou2004learning}, focused on how to effectively propagate labels on a pre-defined graph. The prototype approach is to formulate the problem as regularized function estimation, targeting at a trade-off between the accuracy of the classification function on the labeled points and the regularization that favors a function which is sufficiently smooth with respect to the intrinsic manifold structure revealed by both labeled and unlabeled points. These label propagation methods can effectively spread labels, given the underlying manifold is appropriately represented by the affinity graph. What if the graph itself is problematic? That is indeed the case during the past decade, where a typical graph is constructed using Gaussian kernel in the Euclidean space, as we know Euclidean distance cannot well approximate the geodesic distance on the manifold, especially for high-dimensional data due to the ``curse of dimensionality".

Recently, research focus of GSSL has been shifted to constructing an adequate graph~\cite{li2015learning,zhuang2017label,zhang2018semisupervised,fang2018flexible} that better reveals the data manifiold in order to facilitate the following label propagation. State-of-the-art graph construction methods are usually based on the self-expressive model which reconstructs each data point by a liner combination of all other data points. The reconstruction coefficients, regularized by sparsity or low-rankness, are then used to construct a graph. Despite the success of these methods on many application, a common drawback is that the correlation between the label and graph is not fully exploited, as \textit{graph construction} and \textit{label propagation} are formulated as two independent stages. Typical pipeline of current GSSL methods start with constructing a graph using Gaussian kernel or sparse representation, and then obtain results by applying label propagation, \textit{e.g.}, GFHF or LGC, on the \textit{static} graph defined in the previous stage. Although limited works~\cite{zhuang2017label} attempted to use initial label information to guide the graph construction, it is still suboptimal as it has been shown in~\cite{li2015learning} that even the estimated labels of the unlabeled points can provide ``weakly" supervision for building a better graph.

In this work, we attempt to integrate graph construction and label propagation into one unified iterative framework. Specifically, we formulate graph construction as a regularized function estimation problem similar to label propagation. We show that these two function estimation problems can be solved efficiently by two iterative diffusion processes that are fundamentally similar. Most importantly, by alternately or jointly running these two diffusion processes, we are allowed to fully interact the two stages in a iterative manner, as in each iteration, labels are propagated on a dynamic graph updated simultaneously with the supervision of the estimated labels obtained in the previous iteration.

\textbf{Paper contribution}. The main contribution of this paper can be summarized as three folds:\begin{itemize}
	\item We formulate the graph construction as a regularized optimization problem that shares the same intuition as label propagation. They both attempt to fit the label information while regularizing the function smoothness using graph Laplacian.
	\item We propose an iterative diffusion process that can fully exploit the given and estimated labels to efficiently solve the optimization problem.
	\item We integrate graph construction and label propagation into one unified iterative diffusion framework, allowing them to be updated simultaneously. 
\end{itemize}  
  
\section{Related work}
\subsection{Graph-based semi-supervised learning}
Graph construction is of great importance to the success of all GSSL methods, as the effectiveness of the latter label propagation stage depends heavily on having a accurate graph that well reveals the underlying manifold. Started with a Gaussian kernel $W_{ij}=\exp\left(\nicefrac{-\Vert x_i-x_j\Vert^2}{2\sigma^2}\right)$, where $\mathbf{W}$ is called a weight matrix or an affinity/similarity matrix that is equivalent to a graph, recent works~\cite{he2011nonnegative,zhuang2015constructing} on GSSL have focused on constructing a graph using the self-expressive model, \textit{i.e.}, expressing each data point as linear combination of all other points, while regularizing the coefficients by sparsity or low-rankness. However, these approaches arisen in subspace clustering does not fit into the SSL scenario as the given labels are totally neglect. A nature extension~\cite{zhuang2017label} is to add a hard constraint to force the affinities between points with different labels to be zero. While these label information guided graph construction methods~\cite{zhuang2017label,fang2018flexible} showed a reasonable performance gain, they are still suboptimal as only given labels are exploited once, which often account a small portion. It was shown in~\cite{li2015learning} that the estimated unknown labels can provide additional useful supervision for graph construction, given a proper feedback mechanism.

After a graph is constructed, the next step is to propagate label on it. Many classic GSSL methods~\cite{zhu2002learning,zhu2003semi,zhou2004learning} formulated label propagation as a regularized function estimation problem consisting of two terms $\mathcal{J}_{fit}$ and $\mathcal{J}_{smooth}$, namely \textit{fitness} and \textit{smoothness} respectively. The target classification function $\mathbf{F}\in\mathbb{R}^{n\times c}$ can then be obtained by finding a trade-off between these two terms:
\begin{align}
	\arg\min_{\mathbf{F}}\left(\mathcal{J}_{fit}(\mathbf{F})+\mu\mathcal{J}_{smooth}(\mathbf{F})\right).
\end{align}  
A common choice of the fitness term is a quadratic loss between the predicted labels and the groundtruth labels, while imposing graph Laplacian as a smooth operator for the second term, such as GFHF~\cite{zhu2003semi}:
\begin{equation}\label{gfhf}
\min_{\mathbf{F}}\infty\sum_{i=1}^\ell\left\Vert F_{i\cdot}-Y_{i\cdot}\right\Vert^2+\frac{1}{2}\sum_{i,j=1}^nW_{ij}\left\Vert F_{i\cdot}-F_{j\cdot}\right\Vert^2,
\end{equation}
where $\ell$ is the number of labeled points. Note that the infinity weight $\infty$ clamps the predictions on the labeled points to be the given labels. LGC~\cite{zhou2004learning} relaxed the infinity weight to a trade-off parameter $\mu > 0$, and applied on all data points instead of only labeled ones. GGMC~\cite{wang2013semi} followed the same cost function as LGC but reformulated it to a bivariate optimization problem with respect to not only the classification function $\mathbf{F}$ but also the binary label matrix $\mathbf{Y}$. Updating these two variables alternately, it can be seen as learning $\mathbf{F}$ iteratively by feeding back the estimated labels to reinitialize the given labels $\mathbf{Y}$, resulting in a robust classification function less depended on the given labels. A novel label propagation method based on KL-divergence was also proposed in~\cite{fan2018semi}.

\subsection{Diffusion process}
Diffusion is a mechanism which propagates information through a graph represented by an affinity matrix. It has been applied in many computer vision problems, such as clustering~\cite{li2018affinity}, saliency detection~\cite{lu2014learning}, image segmentation~\cite{yang2013affinity}, low-shot learning~\cite{douze2018low} and mainly image retrieval~\cite{donoser2013diffusion,bai2017regularized,iscen2017efficient}.

The diffusion process can be interpreted as random walks on the graph, where a transition matrix $\mathbf{P}$ obtained by normalizing the initial weight matrix $\mathbf{W}$, \textit{i.e.}, $\mathbf{P}=\mathbf{D}^{-1}\mathbf{W}$, defines the probability of walking from one node to another. If the goal is to learning an affinity matrix $\mathbf{A}$ as in retrieval, the random walk can be formulated as an iterative process:
\begin{equation}
	\mathbf{A}^{(t+1)} = \mathbf{A}^{(t)}\mathbf{P}.
\end{equation}
After $t$ steps, the initial affinity matrix are updated by the probability distribution in $\mathbf{P}^t$. This model was then extended to a popular retrieval system PageRank~\cite{page1999pagerank} by introducing an additional random jump matrix $\mathbf{Y}$:
\begin{equation}
\mathbf{A}^{(t+1)} = \alpha\mathbf{A}^{(t)}\mathbf{P} + (1-\alpha)\mathbf{Y},
\end{equation}
where $0<\alpha<1$ controls the trade-off between walk and jump. ~\cite{yang2013affinity} applied the diffusion on a higher-order tensor graph by the following update mechanism:
\begin{equation}\label{tpg}
\mathbf{A}^{(t+1)} = \alpha\mathbf{P}\mathbf{A}^{(t)}\mathbf{P}^\top + (1-\alpha)\mathbf{I},
\end{equation}
where $\mathbf{I}$ is the identity matrix. It has been shown in the survey paper~\cite{donoser2013diffusion} that this type of diffusion process Eq.(\ref{tpg}) consistently yield better results. ~\cite{bai2017regularized} showed that this diffusion process Eq.(\ref{tpg}) closely related to the regularized optimization problem in GSSL.

While the diffusion process has been successfully used in unsupervised scenario, such as image retrieval, it is not clear how to cope it with label information. ~\cite{li2018srd} attempted to leverage labels to facilitate the affinity learning, but only with given labels, limiting its effectiveness in SSL where few labels are available.
\section{Alternating Diffusion Process}
We consider the problem of learning from labeled and unlabeled points:
\begin{problem}
	(\textbf{Semi-supervised learning}). Given a data point set $\mathcal{X}=\{x_1,...,x_\ell,x_{\ell+1},...,x_n\} \subset \mathbb{R}^d$ and a label set $\mathcal{Y}=\{1,2,...,c\}$, the first $\ell(\ell\ll n)$ points $x_i(i\leq \ell)$ are labeled as $y_i\in \mathcal{Y}$ and the other points $x_u(\ell+1\leq u\leq n)$ are unlabeled. The goal of SSL is to predict the label of the unlabeled points. 
\end{problem}

Let a matrix $\mathbf{F} \in \mathbb{R}^{n\times c}$ denote a classification on the point set $\mathcal{X}$, where its entry $F_{ij}$ represents the probability of $x_i$ belonging to $j$-th class. $\mathbf{F}$ is nonnegative and each row sum up to 1. The target label $y_i$ can be obtained by $y_i=\arg\max_jF_{ij}$. Let us transfer the labels to another $n\times c$ matrix $\mathbf{Y}$ by one-hot-encoding,  where $Y_{ij}=1$ if $y_i=j$ and $Y_{ij}=0$ otherwise. We first show how to construct a graph represented by an affinity matrix $\mathbf{A} \in \mathbb{R}^{n\times n}$.

\subsection{Given classification F, update graph A}
\subsubsection{Optimization problem}
With classification $\mathbf{F}$, we formulate the graph construction as a regularized function estimation problem:
\begin{align}\label{optimization}
\min_{\mathbf{A}} \frac{1}{2}\sum_{i,j,k,l=1}^n W_{ij}W_{kl}\left( \frac{A_{ki}+Z_{ki}}{\sqrt{D_{ii}D_{kk}}} - \frac{A_{lj}+Z_{lj}}{\sqrt{D_{jj}D_{ll}}} \right)^2  \nonumber\\
+ \left(-2\sum_{k,i=1}^{n}A_{ki}Z_{ki}\right) + \mu\sum_{k,i=1}^n\left(A_{ki}-I_{ki} \right)^2,  
\end{align}
where $\mu > 0$ is a regularization parameter, $\mathbf{D}$ is a diagonal matrix with its element $D_{ii}=\sum_{j=1}^nW_{ij}$, $\mathbf{I}$ is the $n\times n$ identity matrix and $\mathbf{Z}=\mathbf{F}\mathbf{F}^\top$ is the label similarity/affinity matrix. A large element $Z_{ij}$ indicates that the label vector $F_{i\cdot}$ is similar to $F_{j\cdot}$.

As shown in Eq.(\ref{optimization}), the objective function consists of three terms. The first term is a local smoothness term, where it encourages nearby points identified by $\mathbf{W}$ taking large values on the learned affinity $\mathbf{A}$. Instead of considering pair-wise similarity independently on the original graph represented by $\mathbf{W}$, ADP attempts to smooth $\mathbf{A}$ with 4 nodes at a time by using a higher-order tensor graph~\cite{yang2013affinity}. Specifically, if $W_{ij}$ is large ($x_i$ is similar to $x_j$) and $W_{kl}$ is large ($x_k$ is similar to $x_l$), then $A_{ki}$ and $A_{lj}$ are encouraged to be similar after adding the label similarity. The second term is a fitness term. It encourages a large $A_{ki}$ when $Z_{ki}$ is large. The assumption is that data points with similar label (large $Z_{ki}$) should have a large similarity as well. The last term is a regularization term controlling the scale of $\mathbf{A}$.

It can be seen that the problem of constructing graph $\mathbf{A}$ Eq.(\ref{optimization}) is quite similar to the problem of label propagation such as Eq.(\ref{gfhf}). They both can be seen as estimating a local smooth function on the graph $\mathbf{W}$, while regularizing the global fitness to the labels.

The objective function Eq.(\ref{optimization}) can be transfered to:
\begin{align}\label{objective}
\mathcal{J} = ~&vec(\mathbf{A+Z})^\top\mathbb{L}vec(\mathbf{A+Z}) \nonumber \\
&- 2vec(\mathbf{A})^\top vec(\mathbf{Z}) + \mu\left\Vert vec(\mathbf{A})-vec(\mathbf{I})\right\Vert^2,
\end{align}    
where $\mathbb{L}=\mathbb{I}-\mathbb{S}$ is the normalized graph Laplacian of the tensor product graph. $\mathbb{S}^{nn\times nn}$ is the Kronecker product $\mathbf{S}\otimes \mathbf{S}$, where $\mathbf{S}=\mathbf{D}^{\nicefrac{-1}{2}}\mathbf{W}\mathbf{D}^{\nicefrac{-1}{2}}$. \(\mathit{vec}:\mathbb{R}^{m\times n}\rightarrow \mathbb{R}^{mn}\) is an operator that stacks the columns of a matrix into a column vector. Its inverse is denoted as \(\mathit{vec}^{-1}\).

Taking the partial derivative of $\mathcal{J}$ with respect to $\mathit{vec}(\mathbf{A})$, we have
\begin{align}\label{derivative}
\frac{\partial\mathcal{J}}{\partial vec(\mathbf{A})} = ~&2\left(\mathbb{I}-\mathbb{S}\right)\left(vec(\mathbf{A})+vec(\mathbf{Z})\right) \nonumber \\
&\quad- 2vec(\mathbf{Z}) + 2\mu\left(vec(\mathbf{A})-vec(\mathbf{I})\right).
\end{align}
Setting this derivative Eq.(\ref{derivative}) to zero, we obtain
\begin{align}\label{vecA}
vec(\mathbf{A}) = ~&\frac{1}{\mu+1}\mathbb{S}\left(\mathbb{I}-\frac{1}{\mu+1}\mathbb{S}\right)^{-1}vec(\mathbf{Z}) \nonumber \\
&\quad + \frac{\mu}{\mu+1}\left(\mathbb{I}-\frac{1}{\mu+1}\mathbb{S}\right)^{-1}vec(\mathbf{I}).
\end{align}
After setting $\alpha=\frac{1}{\mu+1}$ and applying $\mathit{vec}^{-1}$ on both sides of Eq.(\ref{vecA}), the closed-form solution can be obtained as
\begin{align}\label{solution1}
\mathbf{A}^{\ast} = ~&vec^{-1}\left(\left(\left(\mathbb{I}-\alpha\mathbb{S}\right)^{-1}-\mathbb{I}\right)vec\left(\mathbf{Z}\right) \right.\nonumber \\ 
&\quad + \left. \left(1-\alpha\right)\left(\mathbb{I}-\alpha\mathbb{S}\right)^{-1}vec\left(\mathbf{I}\right)\right).
\end{align}
Note that $(\mathbb{I}-\alpha\mathbb{S})^{-1}$ is a diffusion kernel~\cite{zhou2004learning}. Though we can obtain the closed-form solution, it is impractical to use due to the heavy computation of the tensor inverse. Next, we introduce an efficient iterative diffusion process which converges to the same solution as Eq.(\ref{solution1}).

\subsubsection{Iterative solver}
The closed-form solution can be efficiently obtained by running the following iteration:

\begin{equation}\label{iteration}
\mathbf{A}^{(t+1)} = \alpha\mathbf{S}\left(\mathbf{A}^{(t)}+\mathbf{Z}\right)\mathbf{S}^\top+\left(1-\alpha\right)\mathbf{I}. 
\end{equation} 
As shown in Eq.(\ref{iteration}), the affinity matrix $\mathbf{A}$ can be learned by iteratively spreading the previous affinity values with the label similarity, while continuously drawing information from the prior affinity $\mathbf{I}$.

Next, we prove the convergence of the iteration. By applying the operator $\mathit{vec}$ on both sides of Eq.(\ref{iteration}), we obtain
\begin{align}\label{vecAt}
vec\left(\mathbf{A}^{(t+1)}\right) =
~&vec\left(\alpha\mathbf{S}\mathbf{A}^{(t)}\mathbf{S}^\top\right) \nonumber \\
&+ vec\left(\alpha\mathbf{S}\mathbf{Z}\mathbf{S}^\top\right)  + \left(1-\alpha\right)vec\left(\mathbf{I}\right).
\end{align}
As $vec\left(\mathbf{A}\mathbf{B}\mathbf{C}^\top\right)= \left(\mathbf{C}\bigotimes \mathbf{A}\right)vec\left(\mathbf{B}\right)$, Eq.(\ref{vecAt}) can be rewritten to
\begin{align}\label{vecAt1}
vec\left(\mathbf{A}^{(t+1)}\right) =
~&\alpha\mathbb{S}vec\left(\mathbf{A}^{(t)}\right) \nonumber \\
&+\alpha\mathbb{S}vec\left(\mathbf{Z}\right) +
\left(1-\alpha\right)vec\left(\mathbf{I}\right).
\end{align}
By running the iteration for $t$ times, we obtain
\begin{align}\label{vecAt2}
vec\left(\mathbf{A}^{(t+1)}\right) =
~&\left(\alpha\mathbb{S}\right)^tvec\left(\mathbf{A}^{(1)}\right) + \sum\limits_{i=1}\limits^{t}\left(\alpha\mathbb{S}\right)^ivec\left(\mathbf{Z}\right) \nonumber \\
&\qquad + \left(1-\alpha\right)\sum\limits_{i=0}\limits^{t-1}\left(\alpha\mathbb{S}\right)^ivec\left(\mathbf{I}\right).
\end{align}
Since the eigenvalues of $\mathbf{S}$ are bounded in $[-1,1]$ and $0 < \alpha < 1$, we have
\begin{equation}\label{withBounded}
\lim\limits_{t\rightarrow \infty}\left(\alpha\mathbb{S}\right)^t=0, \qquad
\lim\limits_{t\rightarrow \infty}\sum\limits_{i=0}\limits^{t-1}\left(\alpha\mathbb{S}\right)^i=\left(\mathbb{I}-\alpha\mathbb{S}\right)^{-1}.
\end{equation}
Hence, Eq.(\ref{vecAt2}) converges to
\begin{align}\label{limvecAt}
\lim\limits_{t\rightarrow \infty}vec\left(\mathbf{A}^{(t+1)}\right) &=
\alpha\mathbb{S}\left(\mathbb{I}-\alpha\mathbb{S}\right)^{-1}vec\left(\mathbf{Z}\right) \nonumber \\
&+ \left(1-\alpha\right)\left(\mathbb{I}-\alpha\mathbb{S}\right)^{-1}vec\left(\mathbf{I}\right).
\end{align}
After applying $\mathit{vec}^{-1}$ on both sides of Eq.(\ref{limvecAt}), we obtain the solution
\begin{align}\label{solution2}
\mathbf{A}^{\ast} = ~&vec^{-1}\left(\left(\left(\mathbb{I}-\alpha\mathbb{S}\right)^{-1}-\mathbb{I}\right)vec\left(\mathbf{Z}\right) \right.\nonumber \\ 
&\quad + \left. \left(1-\alpha\right)\left(\mathbb{I}-\alpha\mathbb{S}\right)^{-1}vec\left(\mathbf{I}\right)\right),
\end{align}
which is exactly the same as Eq.(\ref{solution1}) obtained by solving the optimization problem Eq.(\ref{optimization}). Note that the solution is independent with the initialization of $\mathbf{A}$. In practice, it is initialized as $\mathbf{S}$ for a faster convergence speed.

\subsection{Given graph A, update classification F}
Once a graph is constructed, the next step is to propagate label on it, where the standard label propagation methods can be directly used, such as GFHF~\cite{zhu2003semi}, LGC~\cite{zhou2004learning}, and GGMC~\cite{wang2013semi}. We take LGC as an example. Given the affinity matrix $\mathbf{A}$ and the initial one-hot label matrix $\mathbf{Y}$, LGC attempts to propagate labels by solving the following optimization problem:
\begin{equation}\label{lgc_optimization}
\min_{\mathbf{F}}\frac{1}{2}\sum_{i,j=1}^nA_{ij}
\left\Vert\frac{F_{i\cdot}}{\sqrt{D_{ii}}}-\frac{F_{j\cdot}}{\sqrt{D_{jj}}}\right\Vert^2 + \frac{\mu}{2}\sum_{i=1}^n \left\Vert F_{i\cdot}-Y_{i\cdot}\right\Vert^2,
\end{equation}
where $\mu > 0$ is a regularization parameter, and $D_{ii}=\sum_{j=1}^nA_{ij}$. Note that Eq.(\ref{lgc_optimization}) and Eq.(\ref{optimization}) share the same intuition. They both attempt to  smooth the target function with normalized graph Laplacian while being regularized by the label information. It is not surprisingly that Eq.(\ref{lgc_optimization}) has an equivalent iterative form:
\begin{equation}\label{lgc_iteration}
\mathbf{F}^{(t+1)} = \alpha\mathbf{S_A}\mathbf{F}^{(t)} + (1-\alpha)\mathbf{Y},
\end{equation}
where $\mathbf{S_A}=\mathbf{D}^{\nicefrac{-1}{2}}\mathbf{A}\mathbf{D}^{\nicefrac{-1}{2}}$. Eq.(\ref{lgc_optimization}) and Eq.(~\ref{lgc_iteration}) converge the the same closed-form solution, $F^\ast=(\mathbf{I}-\alpha\mathbf{S_A})^{-1}\mathbf{Y}$, where $(\mathbf{I}-\alpha\mathbf{S_A})^{-1}$ is another diffusion kernel~\cite{zhou2004learning}.

\subsection{The complete algorithm and its variants}
Until now, we have presented a widely-adopted two-stage GSSL algorithm, where the graph and label are independently optimized by the diffusion process. One of the drawbacks of these two-stage methods is that they do not fully exploit the correlation between the graph and the labels, as it has been shown by~\cite{li2015learning} that the estimated labels can provide ``weakly" supervised information for building a better graph and facilitate label propagation. Unlike the variants of LGC~\cite{zhou2004learning} where the estimated labels are used as an re-initialization to apply another round of LGC, we feed it back a bit further to the graph construction stage to supply addition supervision for capturing the underlying manifold structure. Hence, the proposed algorithm can be seen as an alternating optimization between the affinity matrix $\mathbf{A}$ and the classification function $\mathbf{F}$. The complete algorithm is presented in Algorithm~\ref{alg:adp}.

\begin{algorithm}[htp]
	\caption{Alternating Diffusion Process}
	\label{alg:adp}
	\textbf{Input}: weight matrix $\mathbf{W} \in \mathbb{R}^{n\times n}$, label matrix $\mathbf{Y} \in \mathbb{R}^{n\times k}$ \\
	\textbf{Parameter}: regularizer $\alpha$, threshold $\beta$ \\
	\textbf{Output}: classification $\mathbf{F}$, affinity $\mathbf{A}$
	\begin{algorithmic}[1] 
		\STATE Normalize $\mathbf{W}$ symmetrically $\mathbf{S}\leftarrow\mathbf{D}_{\mathbf{W}}^{\nicefrac{-1}{2}}\mathbf{W}\mathbf{D}_{\mathbf{W}}^{\nicefrac{-1}{2}}$
		\STATE Initialize $\mathbf{F}^{(0)} \leftarrow \mathbf{Y}$, $\mathbf{A}^{(0)} \leftarrow \mathbf{S}$, $t \leftarrow 0$
		\WHILE{$\left\Vert \mathbf{F}^{(t+1)} - \mathbf{F}^{(t)}\right\Vert_F > \beta$}
		\STATE Obtain $\mathbf{F}^{(t+1)}$ by iterating Eq.(\ref{lgc_iteration}) until converge
		\STATE Obtain $\mathbf{A}^{(t+1)}$ by iterating Eq.(\ref{iteration}) until converge

		\ENDWHILE
		\STATE \textbf{return} solution
	\end{algorithmic}
\end{algorithm}

The core of the proposed ADP is that the graph and the classification are optimized \textit{alternately} so that the estimated labels can be fed back to construct a better graph in order to facilitate the label propagation. Motivated by the iterative diffusion processes of these two subproblems, one possible variants of ADP is $\mathbf{F}$ and $\mathbf{A}$ can be optimized \textit{jointly} by running only a single iteration for one variable before updating another, resulting the following updating strategy: \\
\[\left\{
\begin{array}{l}
\mathbf{F}^{(t+1)} = \alpha\mathbf{S}^{(t)}\mathbf{F}^{(t)} + (1-\alpha)\mathbf{Y} \\
\mathbf{A}^{(t+1)} = \alpha\mathbf{S}\left(\mathbf{A}^{(t)}+\mathbf{F}^{(t+1)}{\mathbf{F}^{(t+1)}}^\top\right)\mathbf{S}^\top+\left(1-\alpha\right)\mathbf{I}
\end{array}
\right.
\]
In this formula, instead of running two diffusion processes alternately, one diffusion process consisting of two steps in each iteration is used. It can be seen as propagating labels on a dynamic graph smoothed by the initial Gaussian weight matrix under the supervision of both given and estimated labels. We show empirically this variant, namely ADP1, also works well in the experiment.

\section{Experiments}
\textbf{Experimental setup.} For all GSSL methods, the graph is constructed by adaptive Gaussian sparsed by KNN, where the bandwidth $\sigma$ is set to be the mean distance of 27 nearest neighbors and $k$ is set to 10, unless stated otherwise. Other hyper-parameters are set according to the corresponding authors. For the proposed ADP, $\alpha$ and $\beta$ are set to 0.99 and 1e-2 respectively. All the experiments are repeated 10 times with random chosen labeled points, and the average performances are reported.  

\subsection{Graph construction} 
Graph is the essential component of GSSL. To demonstrate the capability of learning an optimal graph, ADP is compared to the following GSSL methods focused on graph construction: KNN, LLE~\cite{roweis2000nonlinear}, SPG~\cite{he2011nonnegative}, NNLRS~\cite{zhuang2015constructing}, SSNNLRS~\cite{zhuang2017label}, RDP~\cite{bai2017regularized}, FAML~\cite{fang2018flexible}, SRD~\cite{li2018srd}. Most of them use self-expressive model~\cite{elhamifar2013sparse,liu2013robust} to build the graph. Follow the same convention, the initial weight matrix $\mathbf{W}$ in ADP is also obtained by sparse representation~\cite{elhamifar2013sparse}. 

COIL20 and YaleB datasets are used to test them. COIL20~\cite{nene1996columbia} contains 1,400 gray-scale images of 20 objects with resolution 128$\times$128. They are resized to 32$\times$32 and the raw pixel values are used as features. YaleB~\cite{georghiades2001few} consists of 2,414 frontal face images of 38 subjects, with 64 images per subject acquired under various illumination conditions. Following the same convention as in~\cite{fang2018flexible}, the first 15 subjects are used. Each image is cropped and down-sampled to 32$\times$32 and the raw pixel is used as input. Different number of labels per class ($\delta$) are tested.        

\begin{figure}[htp]
	\centering
	\includegraphics[width=0.49\linewidth]{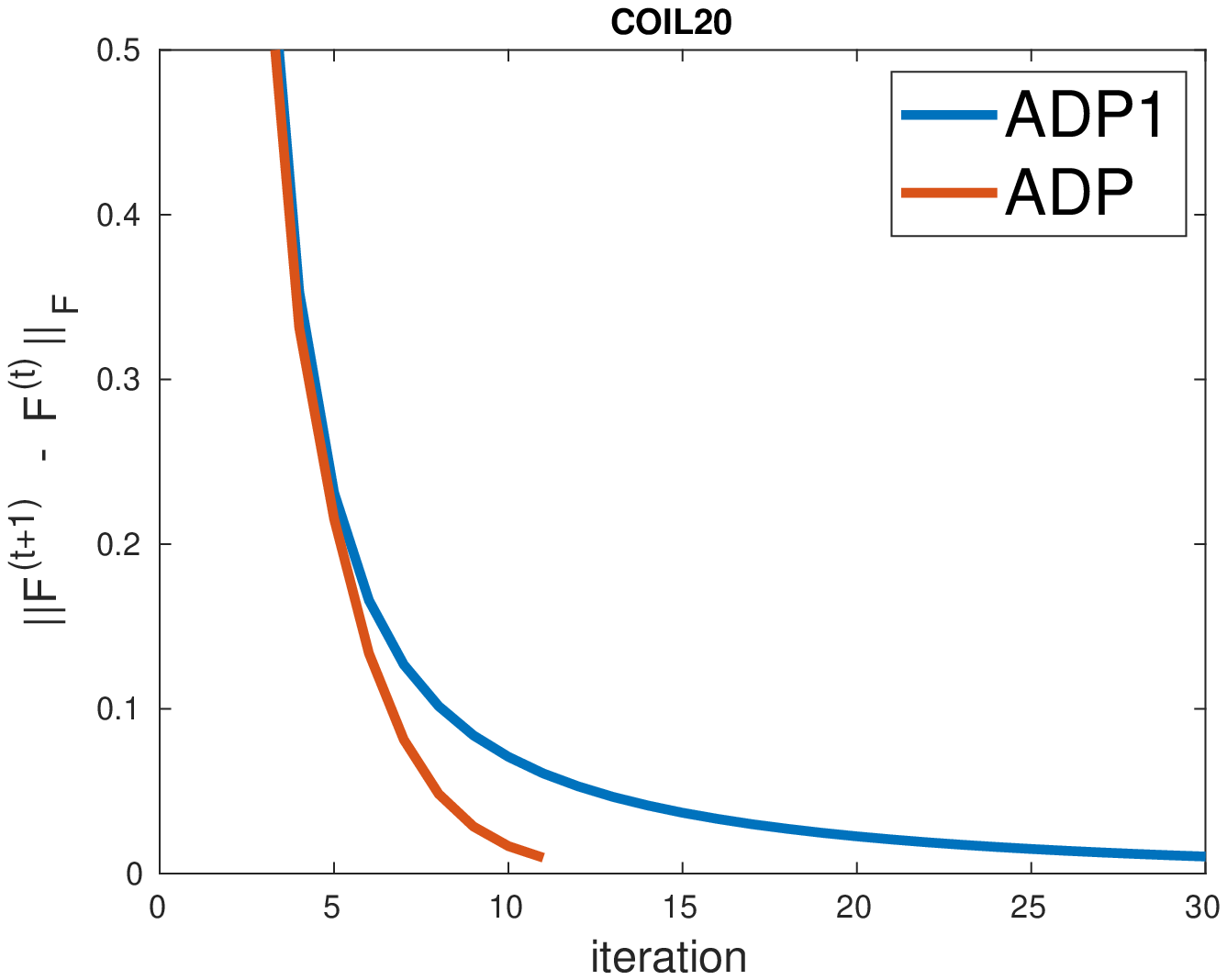}
	\includegraphics[width=0.49\linewidth]{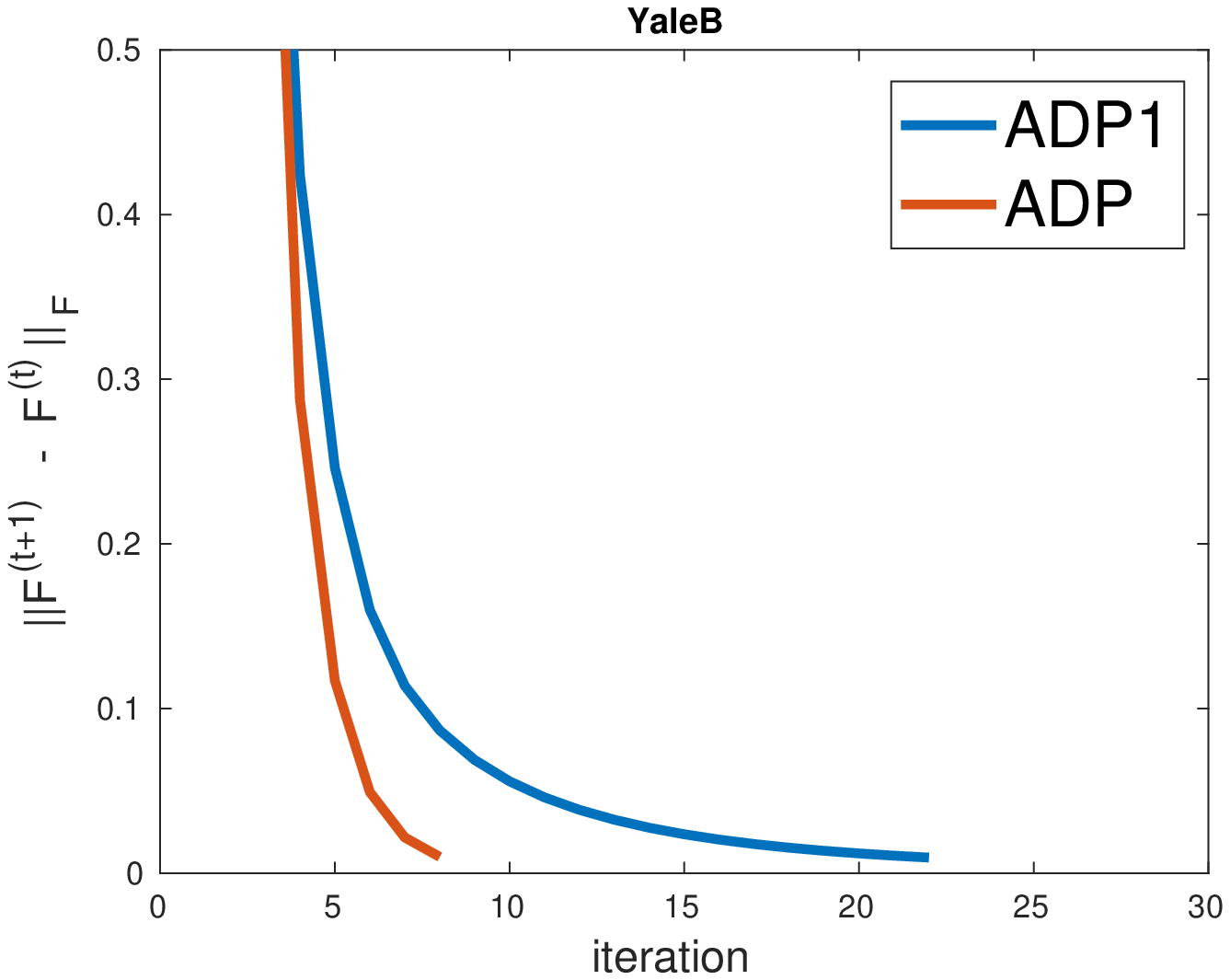}
	\caption{Convergence analysis of the proposed ADP and its variant ADP1 on COIL20 and YaleB datasets.}
	\label{fig:convergence}
\end{figure}

We first check the convergence of ADP and its variant ADP1. As demonstrated in Figure~\ref{fig:convergence}, both ADP and ADP1 converge in several dozens of iterations. Not surprisingly, ADP converges faster than ADP1 as the variables are always updated to its optimal values in ADP, whereas only one iteration is run for each update in ADP1.

Classification results are presented in Table~\ref{tab:graph_construction}. The proposed ADP and ADP1 consistently achieve significantly higher accuracies compared to all other graph construction methods. On the COIL20 dataset, both ADP and ADP1 produce almost perfect results with only three labeled points per class. Similarly, ADP improves 5\% accuracy over the second best method when $\delta = 1$ on the YaleB dataset. Given limited labeled points, the clear advantage of the proposed method comes from the fact that we fully leverage both the given labels and estimated labels to supervise the graph construction, resulting in a robust graph construction strategy less depended on the initial labels. As ADP works slightly better than its variant ADP1, wo focus on ADP in the following experiments. 

\begin{table*}[htp]
	\caption{Classification accuracy (\%) of different graph construction methods with $\delta$ labeled points per class on COIL20 and YaleB datasets. The results marked with (*) are cited from \protect\cite{fang2018flexible}.}
	\label{tab:graph_construction}
	\centering
	\resizebox{\textwidth}{!}{
	\begin{tabular}{ccccccccccc}
		\toprule
		Dataset ($\delta$)   & KNN*     & $\ell_1$-graph*      & SPG*          & NNLRS*         & SSNNLRS*        & RDP            & SRD            & FAML*       & ADP1          & ADP       \\
		\midrule
		COIL20 (1)   & 84.76$\pm$1.75 & 78.01$\pm$3.56 & 77.15$\pm$5.40 & 83.46$\pm$2.62 & 85.02$\pm$3.36 & 87.00$\pm$3.54 & 89.72$\pm$2.80 & 90.54$\pm$2.14 &\textbf{98.54$\pm$1.14}& 96.49$\pm$1.71  \\
		COIL20 (3)   & 91.86$\pm$1.23 & 87.07$\pm$2.32 & 88.62$\pm$4.80 & 91.20$\pm$1.89 & 92.51$\pm$2.11 & 95.00$\pm$1.32 & 96.12$\pm$1.52 & 94.20$\pm$2.18 &99.04$\pm$0.37& \textbf{99.08$\pm$0.55}  \\
		COIL20 (5)   & 93.83$\pm$1.26 & 91.11$\pm$1.73 & 91.88$\pm$3.02 & 93.60$\pm$2.21 & 93.91$\pm$2.05 & 96.81$\pm$1.04 & 97.40$\pm$1.16 & 96.80$\pm$1.41 &\textbf{99.72$\pm$0.25}& 99.44$\pm$0.99  \\
		COIL20 (7)   & 94.89$\pm$1.38 & 93.50$\pm$1.03 & 93.56$\pm$2.50 & 94.16$\pm$2.00 & 94.73$\pm$1.94 & 98.55$\pm$0.37 & 99.01$\pm$0.49 & 97.35$\pm$1.07 &\textbf{99.64$\pm$0.22}& 99.55$\pm$0.42  \\
		COIL20 (9)   & 96.13$\pm$0.69 & 94.46$\pm$0.88 & 94.79$\pm$2.44 & 95.32$\pm$2.06 & 95.89$\pm$1.76 & 98.41$\pm$0.49 & 98.90$\pm$0.52 & 98.34$\pm$0.66 &99.52$\pm$0.13& \textbf{99.58$\pm$0.37}  \\
		COIL20 (11)  & 96.67$\pm$0.44 & 96.00$\pm$0.92 & 95.93$\pm$1.81 & 95.50$\pm$1.75 & 96.21$\pm$1.58 & 98.84$\pm$0.46 & 99.23$\pm$0.47 & 98.49$\pm$0.48 &99.71$\pm$0.09& \textbf{99.80$\pm$0.10}  \\
		Avg.         & 93.02$\pm$1.12 & 90.02$\pm$1.73 & 90.32$\pm$3.34 & 92.20$\pm$2.08 & 93.05$\pm$2.34 & 95.77$\pm$1.20 & 96.73$\pm$1.16 & 95.95$\pm$1.32 &\textbf{99.36$\pm$0.37}& 99.00$\pm$0.69  \\
		\midrule
		YaleB (1)    & 53.40$\pm$3.99 & 47.34$\pm$2.01 & 50.87$\pm$2.69 & 68.96$\pm$3.94 & 75.42$\pm$3.12 & 72.26$\pm$8.46 & 76.26$\pm$8.66 & 86.22$\pm$3.85 &87.54$\pm$4.14& \textbf{91.42$\pm$3.89}  \\
		YaleB (5)    & 72.89$\pm$4.03 & 79.76$\pm$1.35 & 78.20$\pm$3.21 & 84.47$\pm$2.68 & 91.51$\pm$2.39 & 93.44$\pm$0.85 & 94.61$\pm$1.19 & 94.55$\pm$0.98 &94.44$\pm$1.02& \textbf{96.10$\pm$0.74}  \\
		YaleB (9)    & 78.41$\pm$2.43 & 84.90$\pm$1.87 & 88.40$\pm$1.19 & 88.90$\pm$1.87 & 92.91$\pm$2.15 & 95.02$\pm$0.58 & 96.09$\pm$0.77 & 95.88$\pm$1.03 &95.90$\pm$0.89& \textbf{97.37$\pm$0.55}  \\
		YaleB (13)   & 80.13$\pm$1.03 & 86.97$\pm$3.94 & 90.57$\pm$2.55 & 92.66$\pm$2.01 & 93.73$\pm$1.64 & 95.74$\pm$0.37 & 96.65$\pm$0.53 & 96.78$\pm$1.35 &96.52$\pm$0.71& \textbf{97.44$\pm$0.43}  \\
		YaleB (17)   & 82.11$\pm$0.82 & 89.98$\pm$4.08 & 91.37$\pm$1.50 & 93.52$\pm$2.12 & 95.77$\pm$1.47 & 96.47$\pm$0.52 & 97.07$\pm$0.60 & 97.37$\pm$1.10 &96.76$\pm$0.73& \textbf{97.61$\pm$0.45}  \\
		YaleB (21)   & 83.86$\pm$1.71 & 94.43$\pm$1.69 & 93.57$\pm$1.09 & 94.16$\pm$1.19 & 96.21$\pm$1.28 & 96.96$\pm$0.73 & 97.57$\pm$0.51 & \textbf{98.54$\pm$0.60} &97.12$\pm$0.66& 97.85$\pm$0.48  \\
		Avg.         & 75.13$\pm$2.34 & 80.56$\pm$2.46 & 82.16$\pm$2.09 & 87.11$\pm$2.30 & 90.93$\pm$2.01 & 91.65$\pm$1.92 & 93.04$\pm$2.04 & 94.89$\pm$1.48 &94.71$\pm$1.35& \textbf{96.47$\pm$1.09}  \\
		\bottomrule
	\end{tabular}
	}
\end{table*}

\subsection{Label propagation}
Another indispensable component of GSSL is label propagation that spreads the labels to unlabeled points according to the learned graph. To show the proposed ADP can appropriately propagate labels, it is compared to several other label propagation methods, including GFHF~\cite{zhu2003semi}, LGC~\cite{zhou2004learning}, GGMC~\cite{wang2013semi}, LPGMM~\cite{fan2018semi}, on various datasets. 

CMU-PIE~\cite{sim2002cmu} contains more than 40,000 facial images of 68 subjects. We use 20 frontal neutral images of all subjects. The images are cropped and resized to 32$\times$32, and the raw pixel is used as feature. 

Texture25~\cite{lazebnik2005sparse} includes 1,000 texture images from 25 classes. The images are pre-processed by pre-trained VGG-net~\cite{simonyan2014very}, resulting in a 4,096-dimensional feature vector per image.

MPEG7~\cite{latecki2000shape} contains 1,400 silhouette images from 70 classes. The IDSC shape descriptor~\cite{ling2007shape} is used for distance calculation.

\begin{figure}[htp]
	\centering
	\includegraphics[width=0.32\linewidth]{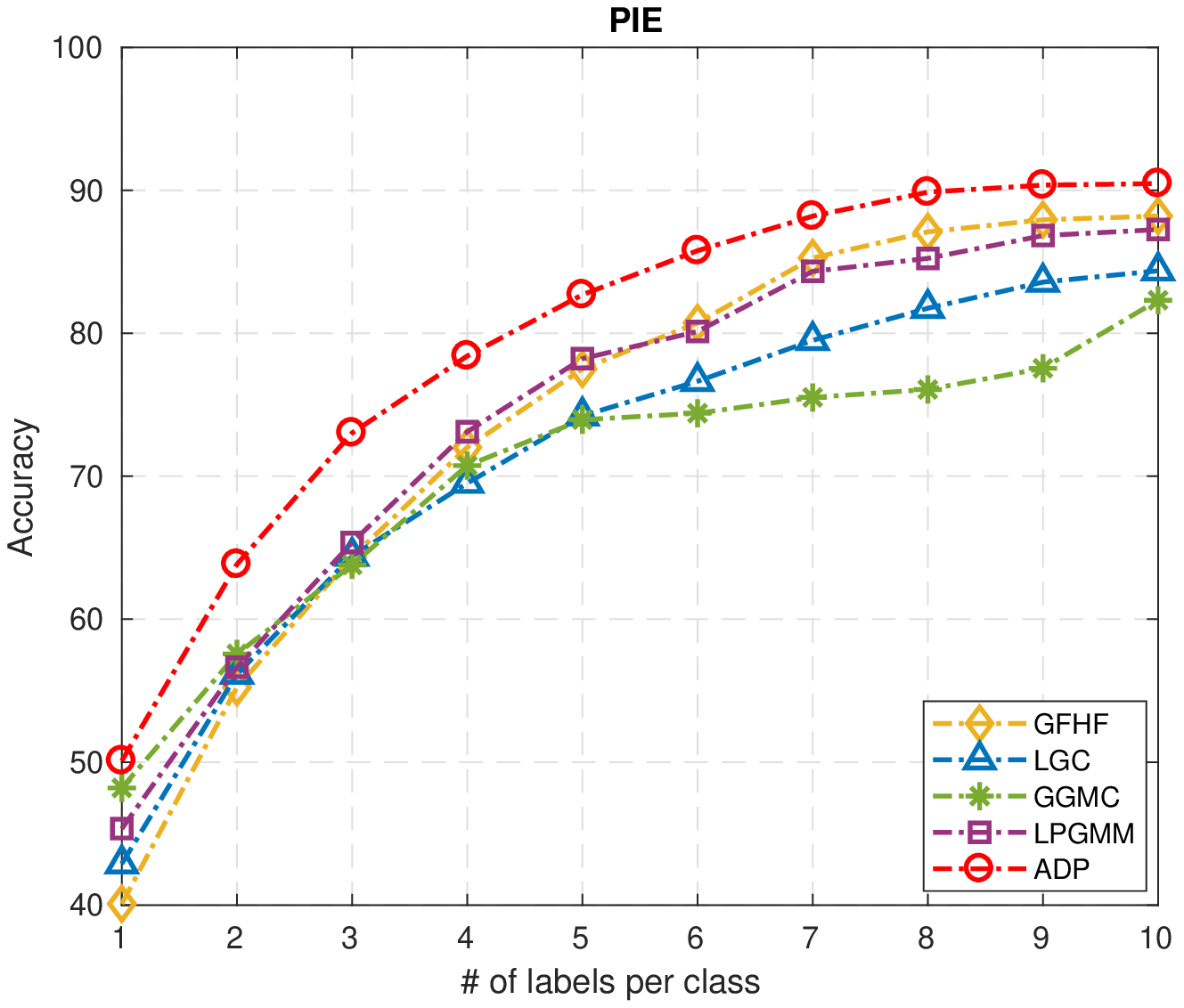}
	\includegraphics[width=0.32\linewidth]{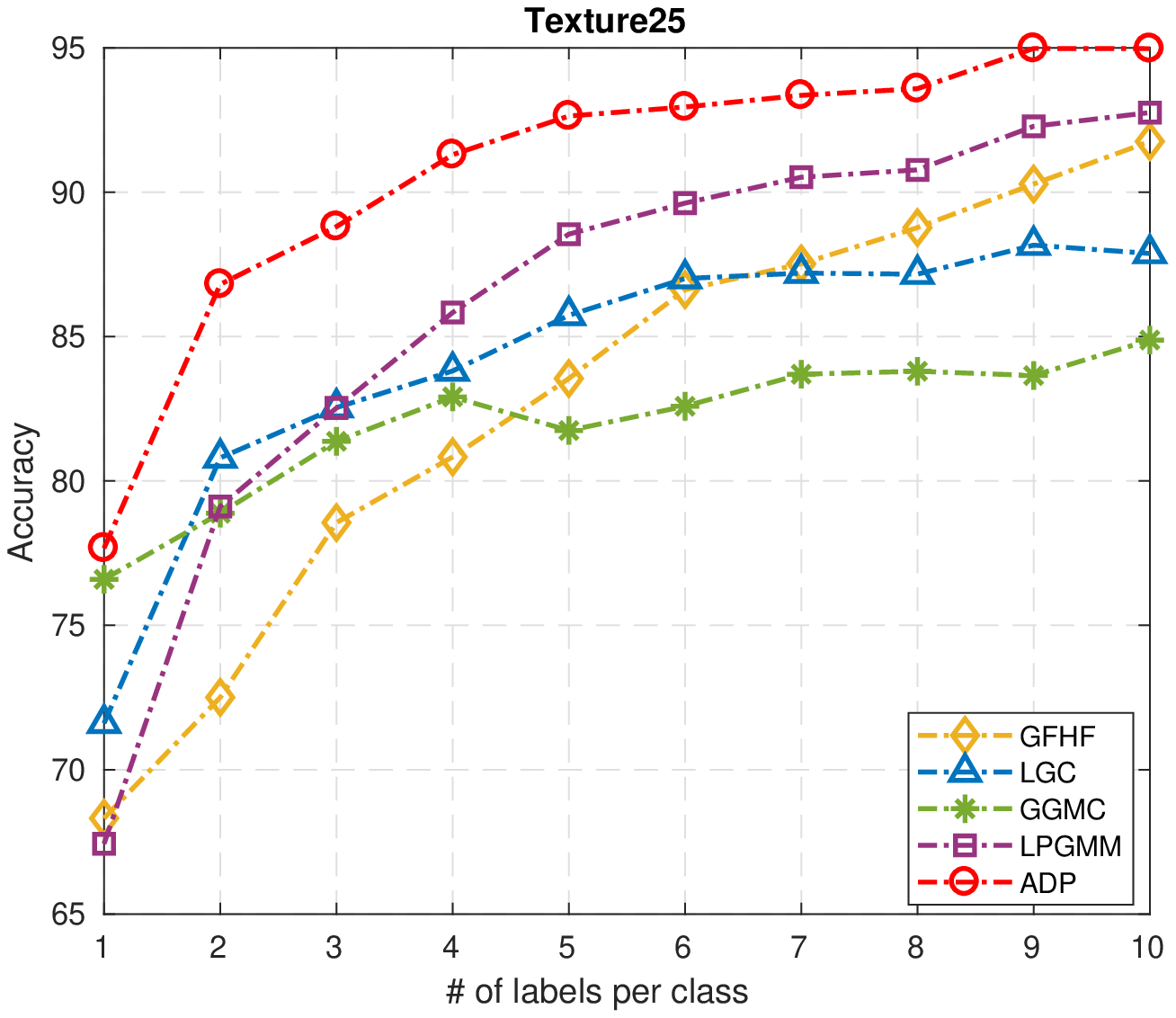}
	\includegraphics[width=0.32\linewidth]{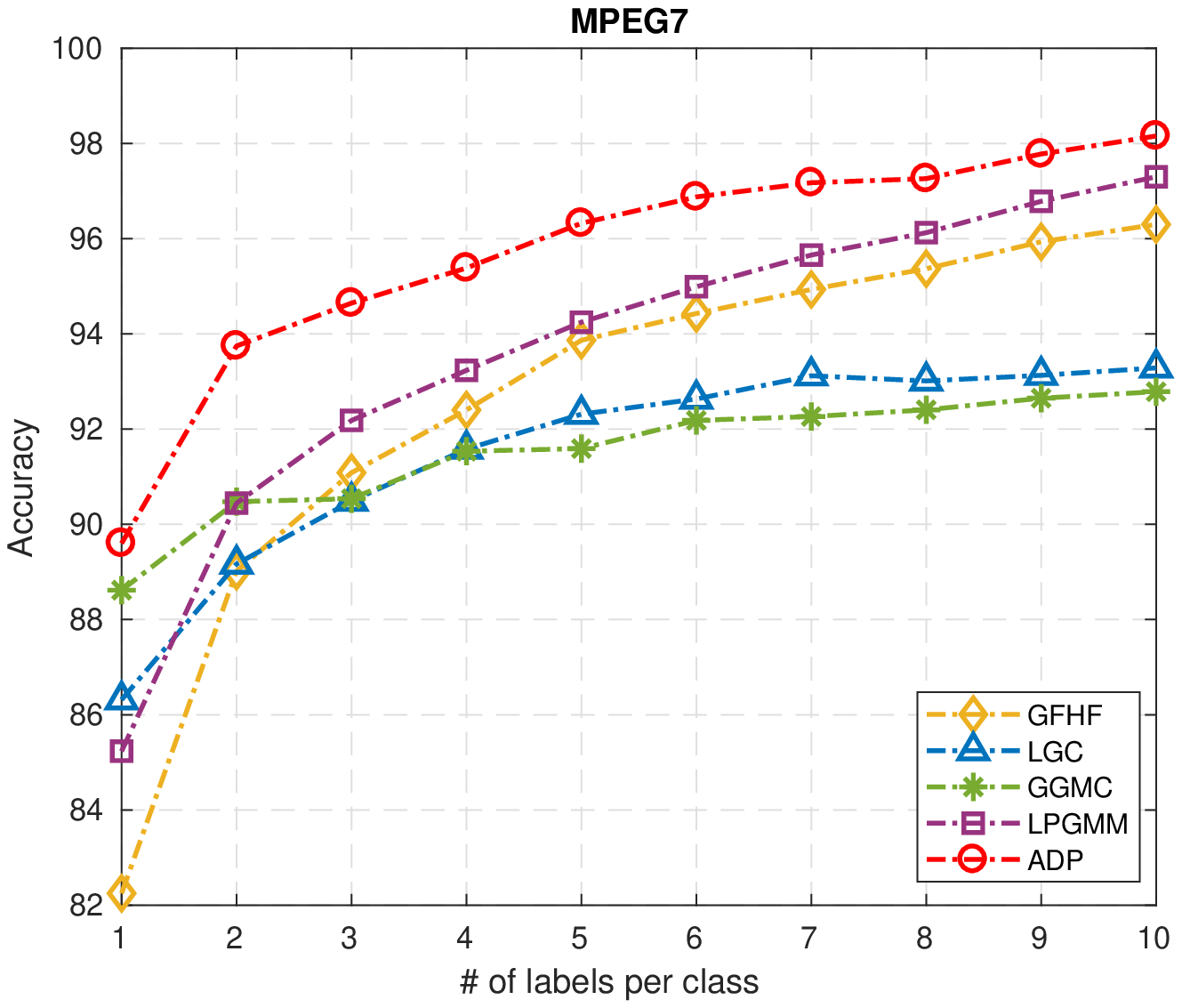}
	\caption{Classification accuracy (\%) of different label propagation methods with $\delta$ labeled points per class on PIE, Texture, and MPEG7 datasets.}
	\label{fig:label_propagation}
\end{figure}

It can be seen from Figure~\ref{fig:label_propagation} that the proposed ADP clearly outperforms other label propagation methods, given different numbers of labeled points on all three datasets. Like the observation in Table\ref{tab:graph_construction}, ADP produces a large winning margin when the labeled points are limited. GGMC has similar effects but it does not gain much improvement when labeled points increase, as shown by~\cite{wang2013semi} as well. All these methods but ADP continuously propagate labels on the initial static graph. This is problematic given a suboptimal graph, and the initial graph is usually suboptimal as few or even no label information is used to construct it. On the contrast, ADP iteratively propagates labels on a dynamic graph updated by the label information obtained in the previous iteration. It guarantees that the labels are propagated with respect to the local manifold structure and the global label information in every iteration.   

\subsection{Compared to other SSL methods}
As shown previously ADP can achieve state-of-the-art performance in GSSL. It is also interesting to see how it compares to other non graph-based approaches, \textit{e.g.}, inductive methods. We select one classic inductive SSL method LapRLS~\cite{belkin2006manifold} as baseline and two state-of-the-art inductive SSL methods, DLSR~\cite{luo2017adaptive} and GD~\cite{gong2018teaching}. Follow the setup of~\cite{gong2018teaching}, we repeat two experiments on the ORL and Wikepedia datasets. It is worth to mention that GD adopted a ``teacher-learner" framework, where a basic SSL learner, holding only the training knowledge, is guided by a teacher with some privileged knowledge. In this experiment, we make only the training knowledge available to ADP. Note that the training performances of these inductive methods are compared, as ADP has no testing phase. Table~\ref{tab:ssl} shows that ADP can also achieve better performance compared to these inductive methods.

\begin{table}[htp]
	\caption{Comparison with other state-of-the-art SSL methods on ORL and Wikipedia datasets. The results marked with (*) are cited from \protect\cite{gong2018teaching}.}
	\label{tab:ssl}
	\centering
	\resizebox{0.95\linewidth}{!}{
		\begin{tabular}{ccccc}
			\toprule
			Dataset          & LapRLS*           & DLSR            & GD*                    & ADP         \\
			\midrule
            ORL       & 0.713$\pm$0.032  & 0.732$\pm$0.012 & 0.776$\pm$0.016 & \textbf{0.804$\pm$0.012}   \\
            Wikipedia & 0.643$\pm$0.010  & -               & 0.663$\pm$0.006 & \textbf{0.675$\pm$0.011}   \\
			\bottomrule
		\end{tabular}
	}
\end{table}

\section{Conclusion}
In this work, we integrated the two separated stages of GSSL, graph construction and label propagation, into one unified framework. We show that learning a graph and a classification can both be formulated as a regularized function estimation problem which attempts to find a trade-off between a global fitness to the labeled points and a local smoothness with respect to the intrinsic manifold structure revealed mainly by the large portion of unlabeled points. An alternating diffusion process was also proposed to efficiently solve the function estimation problems, allowing us to update the graph and unknown labels simultaneously. It is a more plausible solution as we can fully leverage the given labels and unknown labels to construct a local-and-global consistent graph, and also propagate labels in a more effective way thanks to the dynamic graph updated iteratively. Extensive experiments have demonstrated the effectiveness of the proposed method.      
\section*{Acknowledgements}
This work is supported by an Australian Government Research Training Program (RTP) Scholarship.

\bibliographystyle{named}
\bibliography{JDP_IJCAI19}

\end{document}